\documentclass{ifacconf}

\usepackage[T1]{fontenc}
\usepackage[utf8x]{inputenc}

\usepackage{amssymb}
\usepackage{amsmath}

\usepackage{graphicx}      
\graphicspath{ {./figures/} }

\makeatletter
\let\old@ssect\@ssect 
\makeatother

\usepackage{natbib}        
\usepackage[colorlinks=true]{hyperref}
\hypersetup{
    colorlinks=true,
    linkcolor=black,
    filecolor=magenta,      
    urlcolor=blue,
    citecolor=black,
}

\makeatletter
\def\@ssect#1#2#3#4#5#6{%
  \NR@gettitle{#6}
  \old@ssect{#1}{#2}{#3}{#4}{#5}{#6}
}
\makeatother

\usepackage{graphicx}
\begin{document}
\begin{frontmatter}

\title{Modern Machine Learning Tools for Monitoring and Control of Industrial Processes: A Survey}

\thanks[footnoteinfo]{Email:bhushan.gopaluni@ubc.ca\\ \copyright 2020 the authors. This work has been accepted to IFAC World Congress for publication under a Creative Commons Licence CC- BY-NC-ND}

\author[First]{R. Bhushan Gopaluni}
\author[Fifth]{Aditya Tulsyan}
\author[Third]{Benoit Chachuat}
\author[Fifth]{Biao Huang}
\author[Sixth]{Jong Min Lee}
\author[Fifth]{Faraz Amjad}
\author[Fifth]{Seshu Kumar Damarla}
\author[Sixth]{Jong Woo Kim}
\author[Seventh]{Nathan P. Lawrence}

\address[First]{Department of Chemical and Biological Engineering, University of British Columbia, Canada}
\address[Fifth]{Department of Chemical and Materials Engineering, University of Alberta, Canada}
\address[Third]{Department of Chemical Engineering, Imperial College, UK}
\address[Sixth]{Department of Chemical Engineering, Seoul National University, South Korea}
\address[Seventh]{Department of Mathematics, University of British Columbia, Canada}
\begin{abstract}                
Over the last ten years, we have seen a significant increase in industrial data, tremendous improvement in computational power, and major theoretical advances in machine learning. This opens up an opportunity to use modern machine learning tools on large-scale nonlinear monitoring and control problems. This article provides a survey of recent results with applications in the process industry.
\end{abstract}
\end{frontmatter}
\section{Motivation}
Recent advances in Artificial Intelligence (AI) have led to major  breakthroughs in image processing, voice recognition and generation, facial recognition, natural language translation, autonomous driving, and software that can play complex games such as chess and Go \citep{jackson2019introduction, ertel2018introduction}. Currently, there is a tremendous interest in developing AI-based solutions for a variety of longstanding problems in many scientific and engineering disciplines \citep{nilsson2014principles,russell2016artificial}. While some of the promises of AI are potentially hype or misplaced, there are genuine reasons for us to be excited about AI and its impact on process industries \citep{tzafestas2012artificial}. This optimism is driven by the realization that we are at an exciting moment in history with the availability of large volumes of data, enormous computing power, and significant algorithmic advances. Together, we, in the process systems engineering (PSE) community, can develop a vision to design advanced AI-based solutions under the Industry 4.0 paradigm.


We are at the beginning of the fourth industrial revolution that is going to be powered by data \citep{ustundag2017industry, li2017applications}. This revolution is expected to sweep a wide range of industries, including Chemical and Biological industries such as Oil \& Gas, Petrochemicals, Pharmaceuticals, Biotechnology, Pulp \& Paper, and other manufacturing industries. A recent white paper by the consulting firm McKinsey\footnote{\href{https://www.mckinsey.com/~/media/McKinsey/Business\%20Functions/McKinsey\%20Digital/Our\%20Insights/Big\%20data\%20The\%20next\%20frontier\%20for\%20innovation/MGI_big_data_exec_summary.ashx}{www.mckinsey.com}} concluded that the manufacturing industries could generate information worth hundreds of billions of dollars out of existing data and potentially generate much higher value with additional data \citep{brown2011you}. With the advent of the Internet of Things, these industries are poised to generate large volumes of data. This presents an exciting opportunity to exploit data, computers, and algorithms to solve traditional problems surrounding monitoring and control and provide new insights. 

Data analytics and machine learning (ML) ideas are not new to the process industries\footnote{For simplicity; we refer to chemical and biological industries as process industries or just industries}. The review paper by \cite{venkatasubramanian2019promise} provides an excellent overview of the history, the successes and the failures of various attempts, over more than three decades, to use ideas from AI in the industry. In particular, statistical techniques such as principal component analysis (PCA), partial least squares (PLS), canonical variate analysis (CVA), time series methods for modeling such as maximum likelihood estimation and prediction error methods have been extensively used in industry (\cite{chiang2000fault}). Several classification and clustering algorithms, such as k-means, Support Vector Machines (SVMs), Fisher Discriminant Analysis (FDA), are also widely used in the industry (\cite{qin2019advances}). Also, several nonlinear approaches, such as kernel methods, Gaussian Processes (GPs), and adaptive control algorithms, such as Reinforcement Learning (RL), have been applied in some niche applications (\cite{badgwell2018reinforcement,spielberg2019toward,tulsyan2019machine,  tulsyan2019automatic}). Most of these algorithms do not scale well with data, and their performance does not necessarily improve with data.

Through this survey, we are summarizing some of the recent advancements in ML with a focus on its applications in the process industries. We predominantly focus on three application areas -- soft sensing, process monitoring, and process control. For brevity, we only survey four classes of ML methods commonly used in process industries, namely: (1) statistical learning methods; (2) deep learning and its variants; (3) hybrid modeling; and (4) reinforcement learning. Statistical learning methods and deep learning methods are discussed for process modeling, soft-sensing, and process monitoring applications, while hybrid modeling and reinforcement learning are discussed for process control applications. This is by no means an exhaustive survey of the recent research on these topics; however, we have tried our best to include some of the most critical developments of ML tools in process industries. The rest of the article is divided as follows: Section 2 surveys statistical and deep learning methods for soft sensor development in process industries; Section 3 discusses deep learning methods for process modeling and process monitoring; Section 4 surveys reinforcement learning methods for process control applications; Section 5 discusses hybrid modeling techniques for process optimization and control applications; and Section 6 surveys applications of ML tools in applications with limited (or small) data sets.

\section{Statistical and Deep Learning methods for Soft sensor Applications}
Soft sensors have been used in process industries for over two decades to predict product quality and to estimate key process variables (\cite{khatibisepehr2013design,fortuna2007soft}). Soft sensors can be categorized into model-driven and data-driven soft sensors. Model-driven soft sensors (or white box models), such as Kalman filters (and its variants and observers), are based on the first principles models (FPM) that describe the physical and chemical laws that govern a  process (e.g., mass and energy balance equations). In contrast, data-driven soft sensors (or black box models) have no information about the process and are based on empirical observations (historical process data). In between white box and black box models, there is a third type of soft sensors called grey-box models in which a model-driven soft sensor uses a data-driven method to estimate the parameters of an FPM.
\begin{figure}
    \centering
    \includegraphics[width=0.85\linewidth]{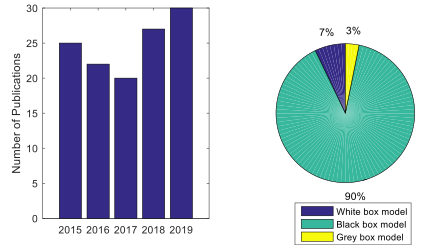}
    \caption{Research publications in soft sensors from 2015 to 2019}
    \label{fig:biao1}
\end{figure}
The statistics in Figure \ref{fig:biao1} indicate that in the last five years, the research in soft-sensing was predominantly focused on data-based models\footnote{Supplementary material explaining the sources of the data used in this section is available at \url{https://dais.chbe.ubc.ca/publications/}}. This is not surprising as data-driven soft sensors are often able to capture complex (and unexplained) process dynamics much more succinctly. In contrast, model-driven soft sensors require a lot of process expert knowledge, which is not always available. Further, model-driven soft sensors are difficult to calibrate, especially for complex nonlinear processes. Note that the grey box model-based soft sensors received the least research attention. The data-driven soft sensors can be further categorized based on the learning technique employed for modeling.
\begin{figure}
    \centering
    \includegraphics[width=0.85\linewidth]{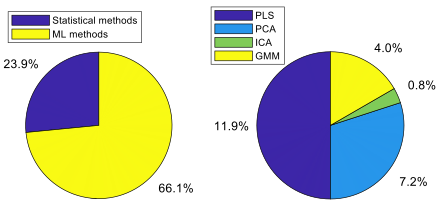}
    \caption{Distribution of data-based modeling techniques (left figure), and statistical methods (right figure)}
    \label{fig:biao2}
\end{figure}
\begin{figure}
    \centering
    \includegraphics[width=0.7\linewidth]{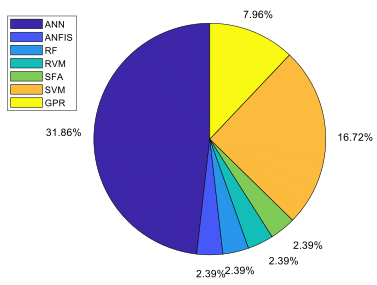}
    \caption{Distribution of ML methods in soft sensing}
    \label{fig:biao3}
\end{figure}
\begin{figure}
    \centering
    \includegraphics[width=0.70\linewidth]{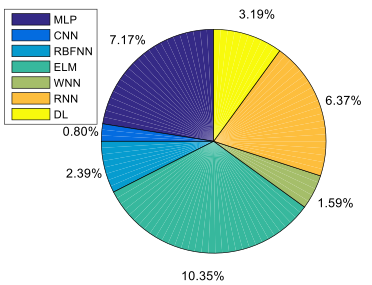}
    \caption{Distribution of various types of ANN in soft sensing}
    \label{fig:biao4}
\end{figure}
\begin{figure}
    \centering
    \includegraphics[width=0.90\linewidth]{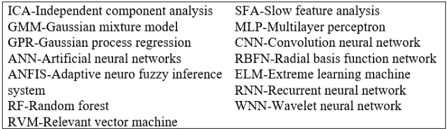}
    \caption{Full forms for acronyms}
    \label{fig:biao5}
\end{figure}

\begin{figure}
    \centering
    \includegraphics[width=0.85\linewidth]{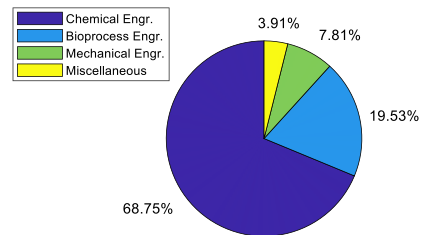}
    \caption{Distribution of soft sensor applications in chemical industries such as refineries, steel plants, polymer industries, cement industries, etc. remain the dominant users of soft sensors.}
    \label{fig:biao6}
\end{figure}

Figures \ref{fig:biao2} to \ref{fig:biao4} show the current trend in the data-driven soft sensing. Figure \ref{fig:biao5} contains full forms for the acronyms used in Figures \ref{fig:biao2} to \ref{fig:biao4}. The research in soft sensing has dramatically shifted from traditional methods to ML methods. Among the ML methods, ANNs received the greatest attention due to their better generalization performance across several different application areas. The class of feed-forward single hidden layer neural networks encompassing MLP, ELM, RBFNN, WNN, etc. has more soft sensing applications than recurrent and deep neural networks. There has been a single research publication reporting the application of Transfer Learning (TL) in inferential measurement. TL may be more suitable for predictive modeling problems that use image data as inputs. TL has not yet been applied for online prediction of process variables. Aside from ANN, SVM is the second most widely used ML method for developing inferential models.

\subsection{Image-based Soft Sensors}

With the development of better camera technology as well as novel image processing techniques, image processing has found increasing use in an industrial setting, sometimes even replacing conventional sensors. There has been extensive research in the application of image processing to the mineral industry (\cite{Horn2017, Poliakov2014, Patel2016, Kistner2013}). In \cite{Horn2017}, for example, online estimation of ore composition on conveyor belts was done. Features and textures were extracted using PCA and Wavelet Texture Analysis (WTA), and then class boundaries were established using SVM. Recently, Convolutional Neural Networks (CNNs) have become popular due to training and hardware breakthroughs, and their use is steadily increasing. In \cite{Horn2017}, CNN was used in the separation of Platinum from other impurities in a froth flotation cell. Images of the froth in the cell were captured using a camera, and CNN was used to extract features from this image. A feature in image processing refers to characteristics, such as edges, shapes, colors, etc., which can be used to distinguish between different objects. Classical ML techniques such as k-NN (k-Nearest Neighbor) and LDA (Linear Discriminant Analysis) were then trained on these features and used to predict Platinum separation. The mineral industry has quickly adapted to using deep learning for image processing (\cite{Fu2018, Carvalho2019, Iglesias2019, Li2019,kholief2017detection}. In the last five years, there have been well over 200 papers on the application of both classical computer vision techniques and CNN in the minerals and ore industry. Computer vision has also been widely used in product fault detection (\cite{Masci2012,Kong2018,Hocenski2016,Rodriguez-Pulido2013}). In \cite{Masci2012}, for example, CNN is trained to detect steel production defects directly on the production line.

Image processing has also found use in the food industry (\cite{Sarkate2013, Siswantoro2013, OBrien2015}). For example, in \cite{Yu2003}, image processing is used to find the amount of coating on chips particles, automating the coating process downstream. Features are extracted from images of chips by first obtaining the 3-channel histogram of the RGB color image, carrying out PCA to reduce the number of features, and then lumping/classifying pixels into ‘L’ classes/bins. Once the features are obtained, PLS regression is used to find the relationship between feature data and the amount of coating on chips. Like the mineral industry, there have been over 200 papers in the field of food quality management in the past five years.
Chemical process industries, on the other hand, have been a little slower in adopting computer vision and deep learning for their applications. This is, however, changing. For example, CNN has been applied to thermal cameras installed in furnaces to get the temperature inside the furnace (\cite{Zhu2018}). A combination of the famous ResNet-50 and U-Net CNN architectures were used for first segmenting (separating/detecting) tube regions in an image from the furnace, and then used to detect the temperature inside the furnace in normal operation. In \cite{Bruno2018}, a CNN is used for classifying results of crystallization trials into four classes (crystals, precipitate, clear, other). The widely used Inception-v3 CNN architecture was used to train on nearly $350,000$ such images, and at test time, it achieved an accuracy of $94\%$. In \cite{Vicente2019}, image processing was used to track the interface level in a vessel with a sight glass in an oil sands process industry. Repeated frame differencing was used to find areas of maximum movement, which corresponds to the interface level. In \cite{Aissa2010}, RGB color analysis is carried out to study and predict powder mixing efficiency. The gist of the procedure is that first, PCA is carried out to map the image's pixels into a lower-dimensional plane. Then secondly, pixel color classification is done using the pixel score scatter plot. Finally, a total count of the number of pixels in each class gives a measure of the area occupied by each color powder. This can be used to calculate the fraction of the area of each powder in a selected region of the image, which gives us a measure of the degree of homogeneity of the powder mixing process. In \cite{eppel2014computer}, computer vision is applied to separate a continuous liquid-liquid flow. Another similar application has been done in \cite{li2018anomaly} in finding the level of a liquid in a process vessel. Overall, there has been comparatively less research in the application of computer vision to process industries, compared to the minerals and food industry, with roughly around $50$ papers in the past $5$ years, mostly for vessel rust detection and pipeline failure, etc.

\section{Deep Learning for Modeling and Fault Detection}
Many recent breakthroughs in computer science and automation can arguably be attributed to a single breakthrough at the heart of all the other advances. That breakthrough is {\em Deep Learning}. Deep Learning refers to Deep Neural Networks (DNN), which are mostly Artificial Neural Networks (ANN) with many layers (typically more than $10$) (\cite{goodfellow2016deep}). A simplistic view of a DNN is that of a mathematical mapping between specific inputs and outputs. As such, at least in theory, it is possible to use DNNs to approximate a wide range of complex nonlinear functions. Universal Approximation Theorem for Feed-Forward ANNs states that a single layer ANN with a sufficiently large number of neurons can approximate a continuous function on a compact subset of $\mathbb{R}^n$ with arbitrary accuracy (with certain mild conditions - please refer to \cite{csaji2001approximation} for technical details). Moreover, with the currently available large-scale optimization tools, such as stochastic gradient descent, it is possible to use DNNs to approximate large-dimensional nonlinear functions. There are two fundamental problems in the process industry where the function approximation power of DNNs is immediately useful: (1) Large-scale nonlinear modeling and (2) Fault Detection and Diagnosis.

\subsection{Large-Scale Modeling}
There are several ways in which models can be used; however, the most common application of DNNs is in the soft-sensing area in industrial processes. One of the first articles on this topic appeared in \cite{shang2014data}. In this article, it was argued that the depth of the network in DNNs provides a high level of flexibility for approximating multi-layered processes. In particular, they demonstrate, on a crude-oil distillation column, that a DNN is superior to other nonlinear modeling approaches. Since this initial work by \cite{shang2014data}, several other authors have pursued this line of research. For instance, in \cite{yan2016data} the performance of a standard DNN was enhanced by combining it with a de-noising auto-encoder. In  \cite{gopakumar2018deep,wang2018deep,villalba2019deep} and the references therein various enhancements of DNN were used to model different processes.

There are different variants of the basic DNN that are more suitable for time-series data common in industrial processes. For instance, standard Recurrent Neural Networks (RNN) and their variant Long-Short-Term Memory (LSTM) networks are two popular categories of DNN. In \cite{wang2018deep}, LSTMs were used to model different phases of a batch process. More recently, there has also been work on using Deep Auto-Encoders (\cite{yuan2018deep}) for soft-sensing. The deep auto-encoders allow us to identify a set of noise-less features that can, in turn, be used for modeling and prediction.

The initial success of deep learning was in image processing. In particular, a variant of DNNs called CNNs had been widely used in image classification, voice recognition, etc. Building on the success of CNNs, a few authors have tried to convert multivariate time-series data into images and make large-scale CNN models for soft-sensing (\cite{wang2019dynamic}). In situations where there is limited data, the idea of transfer learning (where a pre-built DNN is modified slightly) has also been used for soft-sensor modeling (\cite{liu2019domain,patwardhan2019applications}).

DNNs provide an effective nonlinear alternative to traditional soft-sensing methods such as time-series models, PLS, PCA, etc. However, DNNs require large volumes of data with sufficient information in them. As of now, it is not clear what is ``Good Data'' and when it is sufficient to estimate a robust DNN. Moreover, DNNs are black-box models, and hence it is difficult to relate them to the physical properties of a process. However, some recent studies have shown that, despite their complexity, it is possible to study properties such as stability (\citet{bonassi2019lstm}).
\subsection{Fault Detection and Diagnosis}

The flexibility of DNNs in approximating complex nonlinear functions makes them an obvious choice for fault detection and diagnosis. In this context, DNNs have been used in two different ways: (1) as simple classifiers and (2) as tools for generating appropriate lower-dimensional latent space, which in turn is used in process monitoring and fault detection. Recently, several researchers in a wide range of application areas have published work on using DNNs for fault detection and diagnosis.

The vast majority of articles in this area fall under the category of classifiers. The basic idea is to use a DNN classifier to distinguish between normal and faulty operating regions. For instance, in \citet{mireles2018deep} and \citet{lv2016fault}, an RNN was used to build a classifier that identified faults in the well-known Tennessee Eastman benchmark problem. They show that the DNNs perform far better than other commonly used algorithms for fault detection.

The second class of algorithms uses variational auto-encoders to generate a lower-dimensional latent space and use it in conjunction with a statistic to monitor a process and detect any anomalies. An example of this approach is in \citet{wang2019systematic}. More recently, Generative Adversarial Networks have also been used to perform fault detection (\citet{li2018anomaly}).

\section{Reinforcement Learning}


RL refers to the class of numerical methods for the data-driven sequential decision-making problem \citep{Sutton2018}. The RL agent (algorithm) aims to find an optimal policy, or controller, through interactions with its environment. Finding such a policy requires solving the Bellman equation using the principle of optimality; however, it is intractable as it requires a solution in the entire state domain \citep{Bertsekas2005dynamic}. Recent advances in ML enable feature analysis of the raw sensory-level data using deep neural networks (DNNs) and the implementation of various information-theoretic techniques. As a result, Deep RL (DRL) is a general framework that has shown remarkable performances in specific applications such as robotics, autonomous driving, and video games, and board games. \citep{Levine2016, williams2016aggressive, openai2019dota, Silver2017}.

In this paper, we classify algorithms based on the use of model knowledge; that is, model-free and model-based approaches. Model-free is a category of RL in which dynamic models, either first-principle or data-driven forms, are not utilized. Instead, only the reinforcement signal (i.e., reward or cost) is used without model knowledge. The only requirement for underlying system models in the RL framework is a Markov property \citep{puterman2014markov}. Models are often challenging to build, so the ability to bypass the modeling exercise is a considerable advantage for designing the controller. A model-based method, on the other hand, utilizes the state-space model explicitly. Although the process modeling requires extensive domain knowledge, and the results are inaccurate in most cases, it provides a theoretical background of optimality under the specific conditions.

\subsection{Process Control}

With the demands on the performance of process systems, efficient optimization is becoming increasingly essential. The ultimate dream-goal of any process control system will be to develop a controller capable of attaining such optimality in large-scale, nonlinear, and hybrid models with constraints, fast online calculation, and adaptation. This ideal controller should be amenable to a closed-loop solution using a limited number of observations and should be robust to online disturbances. While the mathematical programming (MP) based approaches such as model predictive control and direct optimization have been prevalent, several researchers have also focused on the RL framework, as a complementary method \citep{bucsoniu2018reinforcement, shin2019reinforcement}. This comes from RL's advantages, which allow for designing the feedback controller to interact with its environment and reflect the massive raw data collected throughout the whole industrial process.

Several pioneering pieces of work in the area of PSE applications of RL were first reported in \citep{kaisare2003simulation, lee2006choice, lee2009approximate}. The model-free RL was adopted for solving the optimal control, dual adaptive control, and scheduling problems. It was shown that the approximation of the value function could provide robust control despite the presence of process noise and model changes. These results were later extended to the dynamic optimization and robust optimal control problems \citep{nosair2010min, yang2010probabilistic, yang2013switching}. The recent applications of the RL-based process control approach mainly rely on the linear approximator in both model-based and model-free methods \citep{cui2018factorial, sun2018data, ge2018approximate, kim2018pomdp}. In the meantime, a few DRL applications are studied in \cite{kim2018deep, spielberg2019toward}.

\subsection{Challenges and Advances in DRL}

We highlight several recent advances in the DRL literature with an emphasis on model-free methods. Model-free RL consists of value-based and policy-based methods. In the former, the objective is to find the optimal value function parameter, which minimizes the Bellman error. The policy-based method focuses on the direct optimization of the original objective with respect to the parametrized policy function \citep{deisenroth2013survey}. Classical algorithms for value-based methods, such as Q-learning and SARSA, and policy-based methods, such as REINFORCE, enjoy theoretical convergence. However, convergence can be slow, due to high variance in value estimates, or limited to the tabular setting or linear function approximation \citep{Sutton2018}. Nonetheless, these methods provide the foundation for DRL algorithms.

Deep Q Networks (DQNs) were introduced by \cite{Mnih2015} as a synthesis of deep learning and Q-learning. DQNs are limited to discrete action spaces but showed impressive results in tasks with high-dimensional sensory input data such as Atari games. More recent algorithms inspired by DQNs fall into the actor-critic (AC) framework \citep{konda2000actor}. The goal of AC methods is to jointly optimize the value-based and policy-based objectives. This strategy leverages a parameterized value function to optimize the policy parameters with policy gradient update steps in an online, off-policy fashion. Notably, the Deep Deterministic Policy Gradient (DDPG) algorithm was introduced by \cite{lillicrap2015continuous}. DDPG extends DQN with the use of the deterministic policy gradient theorem \citep{silver2014deterministic}. Despite the advances made by DDPG, it is notoriously difficult to use, for example, due to sensitivity to hyperparameters and overestimation of Q-function values \citep{henderson2018deep, fujimoto2018addressing}. This limits the viability of DDPG for real-world applications such as process control.

More recent DRL algorithms have been geared towards addressing these issues. Twin Delayed DDPG (TD3) addressed the overestimation bias of Q-values as well as variance in target values through the use of two DQNs, delayed actor updates, and target policy smoothing \citep{fujimoto2018addressing}. Concurrently, Soft Actor-Critic (SAC) was developed with similar enhancements from TD3 (such as double DQNs) but is distinct from previous DRL algorithms discussed here \citep{haarnoja2018soft, haarnoja2018applications}. In particular, SAC considers a stochastic actor and optimizes it by maximizing the weighted entropy in an objective similar to the traditional policy-based objective. Such an objective naturally balances \emph{exploitation} and \emph{exploration}.

SAC has been shown to outperform DDPG in sample complexity, robustness, hyperparameters, and performance. In a follow-up paper, \citet{haarnoja2018learning} demonstrate this in benchmark simulated locomotion tasks. These are important measures for any RL algorithm, as a physical system cannot be extensively probed to find suitable hyperparameters. To this end, the authors show that SAC can learn a high-performance policy for a physical quadrupedal robot in around two hours. They demonstrate the robustness of the policy to new terrains as well as to perturbations at the base of the robot.

 Despite these advances, model-free RL algorithms are not sufficiently data-efficient and, therefore, not yet useful in real industrial applications \citep{recht2019tour}. Model-based methods, on the other hand, are often more data-efficient than model-free methods. Several model-based RL algorithms have been developed to solve the Hamilton-Jacobi-Bellman equation adaptively, hence referred to as adaptive DP \citep{Prokhorov1997ACD, Lewis2009RL, jiang2014robust}. The model-based local optimal policy can be parametrized to design the end-to-end controller \citep{Levine2016}. Another extension is to study the connection between the stochastic optimal control framework \citep{theodorou2010generalized, theodorou2010stochastic}. However, it is difficult to build accurate models of industrial processes due to the complexity of these processes and also due to a variety of unknown disturbances \citep{lee2018machine}. There has been recent work focused on unifying model-free and model-based approaches \citep{janner2019trust, du2019task}. This is a useful endeavor, as model-free algorithms can often achieve superior final (asymptotic) performance over model-based approaches, but suffer from relatively weak sample complexity. It is crucial to understand the trade-offs between model-free and model-based methods and their respective data efficiencies from the perspectives of RL and PSE fields. This is an area with tremendous potential for applications that can redefine automation in the process industries.

\section{Hybrid Modeling Methodology and Applications}

In essence, {\em knowledge-driven} (mechanistic) modeling based on first principles and {\em data-driven} modeling constitutes two opposite strategies. Developing a mechanistic model requires a deep understanding of the processes at play and is often labor-intensive. In contrast, a data-driven model requires little physical knowledge, flexible structure, and fast to deploy or maintain. But a larger data set is also typically needed to construct a data-driven model, and its validity may not extend far beyond the conditions under which it was trained. A closely related classification of models concerns their parameterization. The structure of a {\em parametric} model is determined a priori based on mechanistic knowledge, which leads to a fixed number of parameters often with physical or empirical interpretation. By contrast, the nature and number of parameters in a so-called {\em black-box} model are not determined by a priori knowledge but tailored to the data at hand. In between these two extremes lies hybrid {\em semi-parametric} modeling. This approach---also referred to as gray-box or block-oriented modeling in the literature---aims precisely at balancing the pros and cons of pure knowledge- and data-driven methods.

Hybrid modeling has been investigated for over 25 years in chemical and biological process engineering \citep{Psichogios1992, Su1992, Thompson1994, Chen2000}. The claimed benefits of hybrid modeling in these application domains include a faster prediction capability, better extrapolation capability, better calibration properties, easier model life-cycle management, and higher benefit/cost ratio to solve complex problems---see recent survey papers on the development and applications of hybrid models by \citet{VonStosch2014, Solle2017, Zendehboudi2018}. The rest of this section focuses on current trends and challenges in hybrid model formulations and their applications in model-based optimization and control.

\subsection{Hybrid Model Formulations}

A usual classification of hybrid model structures is either as {\em serial} or {\em parallel} \citep{Agarwal1997}. In the serial approach, the data-driven model is most commonly used as an input to the mechanistic model---for instance, a material balance equation with a kinetic rate expressed using a black-box model. This structure is especially suited to situations whereby precise knowledge about some underlying mechanisms is lacking, but sufficient process data exists to infer the corresponding relationship \citep{Psichogios1992, Chen2000}. But when the mechanistic part of the model presents a structural mismatch, one should not expect the serial approach to perform better than a purely mechanistic approach. In the parallel approach, instead, the output of the data-driven model is typically superimposed to that of the mechanistic model \citep{Su1992, Thompson1994}. This structure can significantly improve the prediction accuracy of a mechanistic model when the data-driven component is trained on the residuals between process observations and mechanistic model predictions. Of course, this accuracy may be no better than that of the sole mechanistic model when the process conditions differ drastically from those in the training set.

By far, the most common black-box models used in this context are multilayer perceptrons (MLP) and radial basis function (RBF) based regression techniques \citep{VonStosch2014}. For instance, \citet{Chaffart2018} developed a serial hybrid model of the thin film growth process, which couples a macroscopic gas-phase model described by PDEs to a microscopic thin-film model described by stochastic PDEs via an MLP. But alternative statistical and ML techniques have also been investigated in hybrid modeling. \citet{Ghosh2019} recently proposed subspace identification as the data-driven component in a parallel hybrid model and demonstrated the approach on a batch polymerization reactor. The use of the GP regression, which provides an estimate of the prediction error, is also showing promises in bioprocess engineering applications \citep{Zhang2019b}. Further applications of hybrid modeling in the chemical industry are surveyed by \citet{Schuppert2018}.

The use of parallel hybrid models can significantly alleviate the issue of maintaining or recalibrating a complex mechanistic model by performing the update on the black-box model only. By contrast, training or updating a serial hybrid model is more involved because the outputs of the data-driven component may not be directly measured. Thus, assessing the performance of the black-box model requires simulating the full serial hybrid model and comparing its outputs to the available observations. The classical Akaike Information Criterion or Bayesian Information Criteria can be applied to discriminate among multiple black-box model structures. \citet{Willis2017} recently presented an approach based on sparse regression to simultaneously decide the structure and identify the parameters for a class of rational functions embedded in a serial hybrid model. More generally, one could envision the extension of sparse regression techniques such as ALAMO \citep{Wilson2017} to enable the construction of serial hybrid models.

Looking beyond hybrid semi-parametric models, \citet{venkatasubramanian2019promise} argues for the development of hybrid AI systems that would combine not only mechanistic with data-driven models but also causal model-based explanatory systems or domain-specific knowledge engines. Likewise, the mechanistic model could be replaced by a graph-theoretical model (e.g., signed digraphs) or a production system model (e.g., rule-based representations), creating entirely new fields of research.

\subsection{Model-based Optimization and Control}

A large number of hybrid semi-parametric modeling applications have been for open-loop process optimization. There, a hybrid model is appealing because key operational variables in terms of process performance may be included in the mechanistic part of the model---e.g., to retain sufficient generalization---while capturing other parts of the process using data-driven techniques---e.g., to reduce the computational burden. Either local (gradient-based) or stochastic search techniques have been applied to solve the resulting model-based optimization problems. A notable exception is a work by \citet{Schweidtmann2019} applying complete search techniques to overcome convergence to a local optimum and guarantee global optimality in problems with MLP embedded.

It should be noted that developing a data-driven or hybrid model to speed up the optimization of a more fundamental model is akin to conducting a surrogate-based optimization. The latter constitutes active research in process flowsheet optimization, where existing approaches can be broadly classified into two categories. Global approaches proceed by constructing a surrogate model based on an ensemble of flowsheet simulations before optimizing it, often within an iteration where the surrogate is progressively refined. A number of successful implementations rely on MLP \citep{Henao2011}, GP \citep{Caballero2008,Quirante2015,Kessler2019}, or a combination of various basis functions \citep{Cozad2015,Boukouvala2017} for the surrogate modeling. By contrast, local approaches maintain an accurate representation of the flowsheet (or separate units thereof) within a trust region, whose position and size are adapted iteratively. This procedure entails reconstructing the surrogate model as the trust-region moves around, but it can offer convergence guarantees under mild conditions. Applications of this approach to flowsheet optimization include the work by \citet{Eason2016, Eason2018, Bajaj2018}.

The real-time optimization (RTO) and nonlinear/economic model predictive control (MPC) methodologies use a process model at their core. A majority of the RTO and MPC successful implementations have so far relied on mechanistic models, but there has been a renewal of interest in data-driven approaches; see, for instance, \citet{Lee2018, Yang2019, Wu2019}. In this context, too, hybrid models can play a crucial role in reducing the dependency on data and infusing physical knowledge for better extrapolation capability \citep{Klimasauskas1998, Noor2010}. Recently, \citet{Zhang2019} took the extra step of using the same hybrid model simultaneously in the RTO and MPC layers. Notice that the vast majority of these applications consider serial hybrid models with embedded MLP to approximate complex nonlinearities in the system.

An RTO methodology that exploits the parallel approach of hybrid semi-parametric modeling is modifier adaptation \citep{Chachuat2009}. Unlike classical RTO, modifier adaptation does not adapt the mechanistic model but adds correction terms---the modifiers---to the cost and constraint functions in the optimization model. Original work used the process measurement to estimate linear (gradient-based) corrections \citep{Marchetti2009}. \citet{Gao2016} later explored the use of data-driven approaches based on quadratic surrogates trained with current and past process measurements. More recently, \citet{Ferreira2018} proposed to use GPs as the modifiers for the cost and constraint functions, then \citet{delRio2019} combined this approach with a trust-region algorithm. These developments share many similarities with surrogate-based optimization techniques, with the added complexity that the process data are noisy, and the process optimum might change over time.

\section{ML with limited data}
\begin{figure}[ht]
    \centering
    \includegraphics[width=0.90\linewidth]{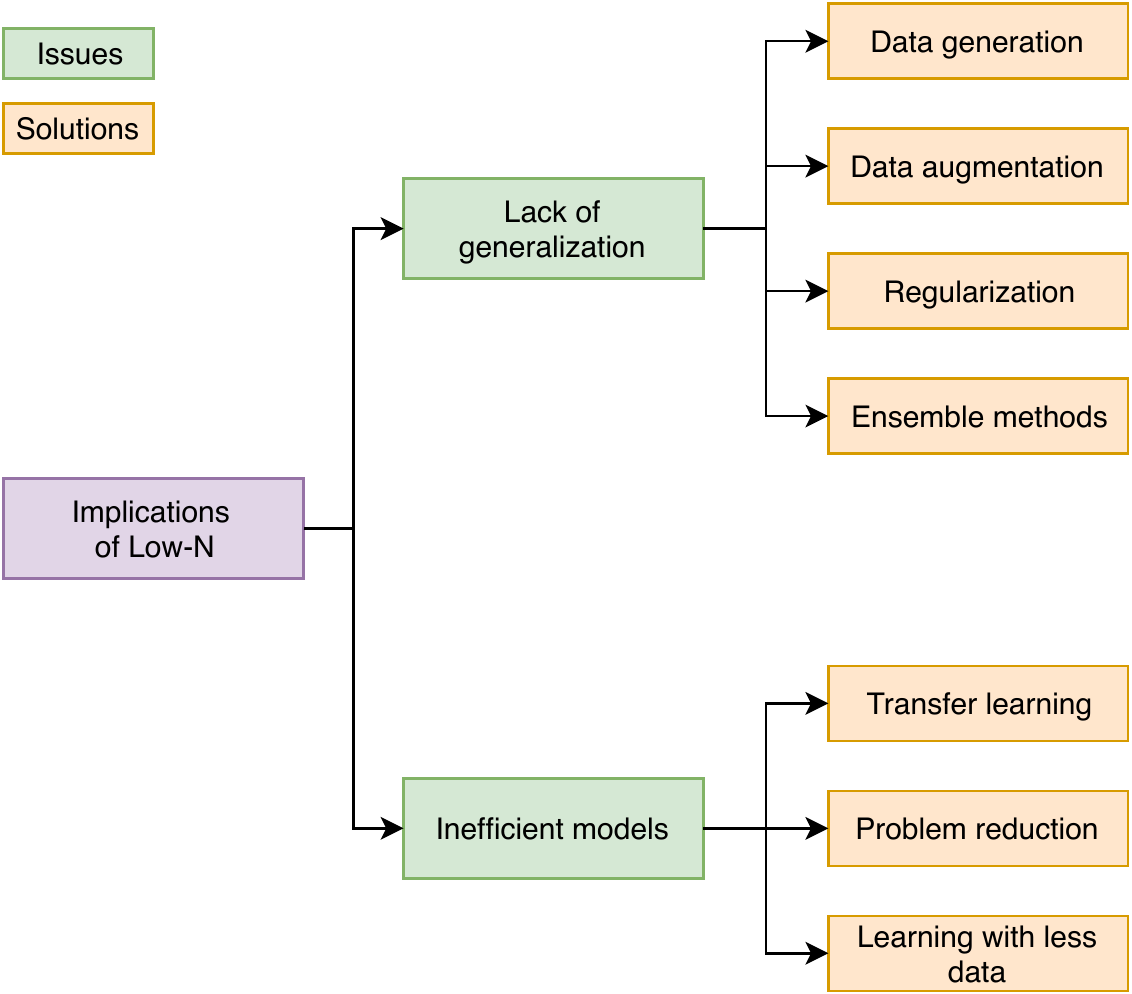}
    \caption{Implications of Low-N in ML \citep{WinNT}.}
    \label{fig:F1}
\end{figure}

The success of ML can be partly attributed to the breakthrough advancements in computation and computing technology and the availability of large amounts of data.  ML continues to make further inroads into sciences (e.g., biomedical, pharmaceutical, and material sciences). However, there is a growing interest in extending the application of ML under Low-N (or small data) settings that define most of these chemical and biological systems. The two critical problems in applying ML under Low-N settings are inefficient model training and model generalization (see Figure \ref{fig:F1}). It is well-known that in the presence of small datasets, ML models (regression or classification) are nontrivial to train and exhibit limited validity. Next, we present a brief survey of traditional and current methods that make ML robust under Low-N settings.

\subsection{Improving Model Generalization}
Several methods based on data generation (e.g., generative models), model regularization (e.g.,  Ridge, LASSO, Elastic Nets), and ensemble methods (e.g., Bagging, Boosting) have been proposed to improve generalization capabilities of ML under Low-N settings.  For brevity, only methods based on data generation are presented here.
Data generators are an essential class of methods that improve model generalization by complementing original small data set with a large number of \emph{in silico} (or computer-generated) data sets.  The idea of data generators was first introduced by \citet{poggio1992recognition} for pattern recognition.  
There are two categories of data generators. The first category of generators extracts prior process knowledge to generate data. For example, \citet{poggio1992recognition} used the 3D view of the object to generate images from any other arbitrary angle. Knowledge-based generators have also been applied in handwritten number recognition \citep{scholkopf1998prior}, text recognition \citep{su2006advances}, and noise source recognition \citep{xu2008research}. In process industries, knowledge-based process simulators and first-principle models have also been extensively used to generate process data under Low-N settings. While knowledge-based generators ensure that the generated samples are representative of the application in-hand, such generators are often nontrivial to construct.

The second class of methods generates \emph{in silico} data by perturbing the original small data set. \citet{lee2000noisy} proposed a perturbation method to generate data by adding a small Gaussian noise to the original samples. Later \citet{wang2008quadratic} introduced a generator that adds a small constant to every dimension of the $n_x$-dimensional training sample, thereby generating $n_x$ \emph{in silico} samples for every training sample.  \citet{gong2006method} then introduced another method that first divides the training samples into $n_x$ groups using $k$-nearest neighbor algorithm, and then generates virtual samples by averaging every two samples in each group keeping the labels unchanged.  Other methods based on probabilistic programming \citep{tulsyan2019industrial,tulsyan2018advances}, Jackknife, and bootstrapping resampling have also been proposed \citep{tsai2008utilize}.
Recently, there has been a growing buzz around a new class of data generators, referred to as generative models. Generative models aim at learning the distribution of the original data set to generate new data points with some variations. Two of the most commonly used and efficient approaches are Variational Autoencoders (VAE) \citep{kingma2013auto} and Generative Adversarial Networks (GAN) \citep{goodfellow2014generative}. Both VAEs and GANs are based on the deep-learning framework. Just in the last five years, several applications of GANs and VAEs have emerged in multiple disciplines encompassing image editing \citep{perarnau2016invertible}, medical image analysis \citep{kazeminia2018gans}, anomaly detection \citep{li2018anomaly}, and financial analysis \citep{wiese2019quant}. Despite the broad appeal of generative models, they require substantially large amounts of training data and also suffer from network instability, which warrants further research. However, there has been some interesting recent work on self-validating classifiers that allow a user to determine if the available limited data is informative enough \citet{}.

\subsection{Improving Model Training}
It is well-known that ML  requires access to large amounts of data for reliable model training  (regression or classification). In the presence of Low-N, model training (i.e., model parameter estimation) becomes a critical issue. Several methods have been proposed to address this problem, including model reduction techniques, novel training methods, and transfer learning. Model reduction techniques primarily aim at reducing the number of model parameters such that the reduced models can be adequately trained with small data sets, often at the cost of model accuracy.

Several attempts have been made to improve model training under Low-N through novel training strategies. For example, \citet{byrne1993generalization} proposed a maximum-likelihood criterion to avoid model over-fitting under the Low-N scenario. A similar Bayesian approach was later introduced by \cite{mao2006new}. Bayesian methods have proven to be useful in identifying robust process models. For example,  \citet{jang2011parameter} proposed a Markov-Chain Monte Carlo (MCMC) method to identify a batch process, while \citet{onisko2001learning} proposed a Bayesian network to identify a manufacturing process.  Similarly,  SVM and neural network (NN) models have also been studied under the Low-N scenario \citep{ingrassia2005neural}. Another approach to improve model training under Low-N is to use adaptive training (or online learning) \citep{zhao2019estimation, barazandegan2015assessment}, where information from new data is incorporated into the model, as and when they become available \citep{zhao2014phase}. For example, in the context of batch process monitoring under Low-N:  \citet{li2000recursive} proposed two adaptive PCA process monitoring algorithms to update the correlation matrix recursively and \citet{flores2004multivariate} extended the multi-block PCA/PLS to incorporate explicitly batch-to-batch trajectory information. \citet{zhao2005double} proposed a double moving window PCA technique for adaptive monitoring, and \citet{lu2004stage} proposed a two-dimensional dynamic PCA, which captures both within-batch and batch-to-batch dynamics simultaneously.

Recently, transfer learning has garnered a lot of interest from researchers as a tool of choice to make model training effective under Low-N settings \citep{pan2009survey}. Transfer learning makes use of the knowledge gained while solving one problem and applying it to a different but related problem. For example, it is now well established that the initial layers of a neural net model, such as ResNet trained on ImageNet data, learn to identify edges and corners in the image, and later layers build on top of these features to learn more complicated structures.  For any new problem whose data looks similar to ImageNet, we can start with pre-trained ImageNet models, change the final layers, and fine-tune it to our dataset. Since the lower layers feature remains relevant, this process makes the optimization process fast and reduces the amount of data required to train new models.  While transfer learning is still an emerging area of research, some of the early successful applications, especially in computer vision tasks \citep{oquab2014learning, shin2016deep}, already make transfer learning an attractive tool under Low-N settings. One step further is meta-learning, or ``learning to learn". Meta-learning algorithms fine-tune some aspect of the underlying learning algorithm; for example, \cite{finn2017model, nichol2018first} develop simple algorithms for any neural network architecture that directly optimizes for initial parameters such that they can quickly be adapted to new tasks with a small amount of data, showing superior performance over standard transfer learning in classification and RL tasks.
\nopagebreak
\section{Conclusions}
There have been several recent successes of modern ML algorithms in a variety of engineering and non-engineering areas. Several of these successes can potentially be ported to process industries. This gives us renewed optimism for developing a new level of enhanced automation in the process industries. We believe that these successes can be translated to build better models, better monitoring and fault diagnosis algorithms, and better controllers using a multitude of data types that include time-series data, text messages, images, and videos. In this survey, we made an effort to provide a vision of this emerging future.
\section{Acknowledgements} We are truly grateful to the anonymous reviewers for their detailed and constructive feedback; however, we apologize for not being able to incorporate all of their suggestions.
\footnotesize
\bibliography{main}             
\end{document}